\documentclass[conference]{IEEEtran}
\IEEEoverridecommandlockouts
\usepackage{amsmath,amssymb,amsfonts}
\usepackage{pdfpages}
\usepackage{algorithmic}
\usepackage{graphicx}
\graphicspath{{plots/}}
\usepackage{textcomp}
\usepackage{xcolor}
\usepackage{float}
\usepackage{booktabs}
\usepackage{pifont}
\usepackage{multirow}
\usepackage[
    backend=biber,
    style=ieee,
    sorting=none,
    natbib=true,
    giveninits=true,
    maxcitenames=2,
    mincitenames=1,
    maxbibnames=99,
    uniquename=false,
    uniquelist=false,
    urldate=long
    ]{biblatex}
\usepackage[hidelinks]{hyperref}

\def\BibTeX{{\rm B\kern-.05em{\sc i\kern-.025em b}\kern-.08em
    T\kern-.1667em\lower.7ex\hbox{E}\kern-.125emX}}
\begin{document}

\title{Preserving Safety During Sequential Fine-Tuning: A Factorial and
    Geometric Analysis of Replay\thanks{Code and results can be found at: \url{https://github.com/gruellan/preserving_safety_sequential_finetuning}}}
\author{\IEEEauthorblockN{George Ruellan}
\IEEEauthorblockA{\textit{School of Electronic Engineering and Computer Science} \\
\textit{Queen Mary University of London}\\
London, United Kingdom \\
g.ruellan@se24.qmul.ac.uk}
\and
\IEEEauthorblockN{Diana Benavides-Prado}
\IEEEauthorblockA{\textit{School of Electronic Engineering and Computer Science} \\
\textit{ Queen Mary University of London}\\
London, United Kingdom \\
d.benavidesprado@qmul.ac.uk}
}
\maketitle

\begin{abstract}
Benign fine-tuning can erode safety alignment in large language models, and sequential adaptation substantially compounds this risk. Prior work mitigates this erosion by replaying harmful-prompt refusals and benign-compliance examples, while other work instead matches base-model logits via distillation. However, the individual contributions of supervised exposure and distillation to safety behavior and refusal geometry during sequential adaptation remain unclear. We factor these two strategies apart and study their behavioral and geometric effects during sequential LoRA fine-tuning of Mistral-7B and Llama-3-8B across diverse task sequences (code generation, German-language summarization, and science question-answering). Behaviorally, supervised replay exposure yields the largest reduction in HarmBench attack success rate (ASR) on Mistral, averaging 51.1 points below base, but at a substantial over-refusal cost, averaging 53.4 points above base on OR-Bench-hard. Logit distillation, without any supervised replay loss, largely prevents safety erosion at a fraction of this cost: an average 5.4-point ASR reduction against only a 5.1-point rise in over-refusal. Llama shows the same asymmetry between the two strategies. Combining exposure and distillation moderates exposure's over-refusal while leaving downstream task scores largely unaffected. Geometrically, the two models diverge: on Mistral, distillation preserves the base refusal direction more closely than exposure does, whereas on Llama, refusal behavior after exposure largely survives ablation of the extracted refusal direction. This dissociation between models indicates that preserving the base refusal direction is not a reliable proxy for behavioral safety. 
\end{abstract}

\section{Introduction}
Fine-tuning can compromise safety alignment even when the downstream data are benign \citep{qi_2024_finetuning, he_2024_whats}. Repeated adaptation makes this a continual learning problem: a model must acquire new task behavior while retaining earlier capabilities and safety constraints \citep{kirkpatrick_2017_overcoming, shi_2025_continual}. LoRA is a fine-tuning technique that learns low-rank weight updates efficiently \citep{hu_2022_lora}. However, it does not guarantee retention of safety \citep{lermen_2024_lora}. A mitigation must therefore limit harmful compliance across successive tasks while preserving the model's ability to answer benign requests.

Replay revisits a subset of past data during adaptation to limit catastrophic forgetting \citep{buzzega_2020_dark}. It has been successfully applied to continual fine-tuning of language models, helping to retain earlier task performance while learning new tasks \citep{scialom_2022_finetuned}. In safety alignment, mixing supervised safety examples into fine-tuning data can reduce harmful responses \citep{qi_2024_finetuning} but excessive safety replay can also increase refusal of benign requests \citep{bianchi_2024_safety_tuned}. Unforgotten Safety \citep{alssum_2025_unforgotten} shows that Dark Experience Replay (DER) \citep{buzzega_2020_dark} reduces harmful compliance after one downstream task while retaining task utility, although over-refusal increases relative to standard fine-tuning. SafeAnchor \citep{guo2026safeanchor} studies sequential domain adaptation across six domain orderings, combining replay with projection and distribution anchoring.

Nevertheless, replay can help achieve safety preservation via two very distinct training signals. Supervised exposure teaches supplied refusal and benign-compliance responses; logit distillation encourages retention of the base model's output distribution. This distinction matters because the base model may already comply with harmful requests or refuse benign ones. Consequently, matching the base model's logits and training on supplied safe responses may produce different safety outcomes from the same replay samples. Furthermore, supervised replay exposure and base-model logit distillation may differ in their behavioral and geometric effects across tasks that vary in format, domain and language, an aspect that has been overlooked in previous studies. Prior work links refusal to early output tokens \citep{qi_2025_safety} and low-dimensional activation and weight representations \citep{arditi_2024_refusal, wei_2024_assessing}. Yet, it remains unclear whether preserving the base model's refusal representation corresponds to safer behavior after fine-tuning.

We decouple supervised replay exposure and logit distillation during sequential fine-tuning on code generation, German-language summarization and science question-answering. We evaluate the effects of having these two signals, having each of these separately and having neither (i.e. standard fine-tuning). Joint measurement of harmful compliance, benign refusal and task performance establishes what each training objective preserves and at what cost. We then track refusal-related activations and use directional ablation to test whether the extracted representation continues to mediate refusal. Our main contributions and findings are:

\begin{itemize}
    \item \textbf{Safety degradation depends on task order.} Across six task orderings, task-only fine-tuning ends with higher attack success rate (ASR) compared to the base model on both Mistral and Llama models, with much of the variation following the final task. This motivates evaluation across whole diverse sequences rather than single tasks.
    \item \textbf{Exposure reduces harmful compliance, while distillation retains base behavior.} Our factorial comparison shows that supervised exposure produces the largest ASR reductions below base and the greatest over-refusal, while distillation changes behavior the least. Combining the signals reduces exposure's over-refusal, but also gives up some of its ASR reduction on a Mistral model. The objectives therefore make distinct, non-additive contributions.
    \item \textbf{Task scores stay similar despite increased over-refusal.} Combined replay lowers ASR below base across six task orderings while retaining similar task scores, yet its benign-refusal cost depends on the model and benchmark. This is a direct observation from joint evaluation that task performance alone cannot capture.
    \item \textbf{Lower ASR does not require preserving the base refusal direction.} By relating geometric trajectories to behavioral outcomes, we show that exposure achieves lower ASR than distillation despite greater rotation of the refusal direction. On Llama, refusal also persists after ablation of the checkpoint's own extracted direction, limiting what a single direction at a fixed layer explains about refusal after replay.
\end{itemize}

The paper is organized as follows. Section~\ref{sec:related_work} reviews related work. Section~\ref{sec:methodology} defines training, evaluation and geometric analysis protocols. Section~\ref{sec:results} presents the sequential erosion problem, separates the behavioral effects of replay objectives and tests what task performance and refusal geometry reveal about those effects, including seed and Gemma replications, and a recent projection method. Sections~\ref{sec:discussion}--\ref{sec:conclusion} discuss the implications, limitations and conclusions of this work.

\section{Background and Related Work}
\label{sec:related_work}

\subsection{Continual Learning and Catastrophic Forgetting}
Following pre-training and alignment, LLMs are often adapted sequentially across domains or tasks, a process called continual learning (CL). \citet{shi_2025_continual} distinguish vertical CL, which repeatedly adapts a general model to one task or domain, from horizontal CL, which adapts it across a sequence of tasks or time periods. We study the latter, also known as sequential fine-tuning, where catastrophic forgetting is a particular risk \citep{luo_2025_an}. Classical mitigations include regularization-based, replay-based and model-based methods \citep{parisi_2019_continual}. Their ability to preserve safety alignment has received less attention than their effect on task retention.

\subsection{The Fragility of Safety Alignment}
\citet{qi_2024_finetuning} show that benign fine-tuning can compromise aligned instruction-tuned models, a finding replicated by \citet{fraser_2025_finetuning} for the Mistral-7B family used here. \citet{qi_2025_safety} attribute this fragility to alignment being concentrated in the first few output tokens, where subsequent fine-tuning can readily alter the model's distribution. In a two-task sequence, \citet{ung2024chained} find greater safety degradation when task fine-tuning follows safety fine-tuning than under the reverse order. We extend this setting to all six permutations of three heterogeneous downstream tasks.

\subsection{Safety Degradation Mitigation}
Relevant mitigations include projection-based, constraint-based and replay-based methods. SafeLoRA \citep{hsu_2024_safe} projects LoRA updates onto a safety-aligned subspace, SaLoRA \citep{li_2025_salora} freezes a safety module during fine-tuning and ProCon \citep{du2025anchoring} penalizes hidden-state drift along the refusal direction. Replay supplements task data with examples of safe behavior during fine-tuning. These approaches target different parts of adaptation: weight updates, hidden representations or model responses.

Existing studies examine both their behavioral outcomes and internal effects. Unforgotten Safety measures over-refusal using XSTest, OR-Bench-80K and OR-Bench-Hard alongside ASR and task utility \citep{alssum_2025_unforgotten}. SafeAnchor combines projection, distribution anchoring and triggered replay and also studies weight-space subspace drift \citep{guo2026safeanchor}. We focus on the separate effects of supervised exposure and logit distillation on behavior and refusal geometry during heterogeneous sequential adaptation, with a SafeLoRA comparison of safety and over-refusal under weight-space projection.

\subsection{Refusal Geometry and Causal Interventions}
Research on model internals has linked refusal to activation-space representations. \citet{arditi_2024_refusal} found that a single vector mediates refusal across 13 open-source models ranging from 1.8B to 72B parameters. Ablating this direction nearly eliminates refusal of harmful prompts, while adding it induces refusal. Other work locates safety in a small band of middle layers \citep{li_2025_safetylayers} or roughly $5\%$ of neurons \citep{chen_2025_neurons}. The single-direction account remains contested: \citet{wollschlger_2025_the} describe a multidimensional concept cone, \citet{pan_2025_orthog} identify multiple orthogonal safety directions and \citet{faaizjoad_2026_there} find a representation more distributed than any single vector. ProCon \citep{du2025anchoring} analyzes refusal-direction drift during fine-tuning, while \citet{lan2026dynamic} tracks behavioral and geometric changes under dynamic adversarial fine-tuning. These findings motivate two separate tests after replay: whether similarity to the base direction tracks harmful compliance, and whether the checkpoint's extracted direction still mediates refusal after adaptation.

\section{Methodology}
\label{sec:methodology}

Our study analyses the effect of the replay objective and the effect of encountering replay examples. We hold the buffer fixed while enabling supervised exposure, logit distillation, both or neither. Behavioral evaluation measures harmful compliance, benign refusal and task learning. We also track each model's refusal direction relative to the base model and ablate the extracted direction to test its role in refusal after adaptation.

\subsection{Training}
\label{sec:training}
We compare task-only fine-tuning (Plain), Experience Replay (ER; supervised replay exposure only), Dark Experience Replay (DER; base-model logit distillation only) and DER++ (both replay signals). A fifth condition, Matched Plain, applies task loss alone using the replay trainer. All these alternatives share adapter and optimizer hyperparameters. Matched Plain, ER, DER and DER++ also share the replay data path, batching and update budget; original Plain uses length grouping and fewer updates. Matched Plain measures the size of this procedural difference (see Appendix~\ref{app:training_budget}).

The main experiments fine-tune LoRA adapters sequentially on three TRACE tasks \citep{wang_2023_trace}: Py150 (code generation), 20Minuten (German-language summarization) and ScienceQA (scientific question-answering). Adapters use rank 16 on the $q/k/v/o$ attention projections, with scaling 32 and dropout $0.05$. Rank 16 is used in prior LoRA safety fine-tuning of Mistral-7B-Instruct and Llama-3-8B-Instruct \citep{xie_2025_deepref} and in sequential safety preservation \citep{guo2026safeanchor} (see Appendix~\ref{app:impl} for other training settings).

The core $2\times2$ comparison covers task orderings O1--O3 on Mistral and Llama, while Plain and DER++ cover all six task orderings, as shown in Table~\ref{tab:ordering_defs}. Supplementary analyses examine a homogeneous task sequence, SafeLoRA projection and Gemma (see Appendix~\ref{app:coverage} for evaluation coverage).

LoRA adapters are merged cumulatively. After task $k$, $M_k$ consists of the base weights $M_0$ plus adapters $1,\ldots,k$, each trained as a delta from the preceding cumulative model. To assess seed sensitivity, we replicate the four core conditions (Section~\ref{sec:er_ft}) at O1 with seeds 42, 43 and 44 on both models. We also train each model on a homogeneous sequence of three Py150 shards (1667/1667/1666 samples) using the same optimizer and adapter settings. The smaller data volume prevents a controlled comparison of homogeneous and heterogeneous task sequences.

\begin{table}[ht]
\centering
\footnotesize
\setlength{\tabcolsep}{8pt}
\caption{Task sequences for ordering labels O1--O6. Code denotes Py150, sum denotes 20Minuten and qa denotes ScienceQA.}
\label{tab:ordering_defs}
\begin{tabular}{lccc}
\toprule
Ordering & Task 1 & Task 2 & Task 3 \\
\midrule
O1 & code & sum & qa \\
O2 & qa & sum & code \\
O3 & sum & code & qa \\
O4 & code & qa & sum \\
O5 & sum & qa & code \\
O6 & qa & code & sum \\
\bottomrule
\end{tabular}
\end{table}

\subsection{Factorial decomposition of replay}
\label{sec:er_ft}

We independently enable supervised replay exposure and base-model logit distillation to distinguish their individual and combined effects. During each task, the full fixed buffer of 500 examples is concatenated with the 5000 task examples and shuffled. Replay therefore accounts for 500/5500 examples per epoch; the number of replay examples in a minibatch varies. We hold buffer size fixed and do not evaluate alternative sizes. The buffer is constructed from WildJailbreak \citep{jiang_2024_wildteaming}: harmful samples pair a \texttt{vanilla\_harmful} prompt with a safe refusal completion and over-refusal samples pair an \texttt{adversarial\_benign} prompt with a compliant completion.

The main buffer contains 350 harmful and 150 benign prompts (70:30). Its builder filters candidates using $M_0$ behavior: harmful prompts must elicit a refusal without harmful content and benign prompts a compliant response. The supervised target remains the WildJailbreak \texttt{completion} field; the separately stored \texttt{m0\_response} records filtering provenance and is not the replay CE target. Teacher logits are computed under the stored completion. The released builder and fixed seed make the buffer composition and any overlap with evaluation sets auditable. Exact regeneration is limited by the unpinned base-model revision used for behavior filtering.

The loss configurations use task examples $X$ and replay examples $X_r$ with the following terms:
\begin{equation}
\begin{aligned}
L ={}& CE(f_{\theta}(X), Y)
      + \beta \cdot CE(f_{\theta}(X_r), Y_r) \\
     &+ \alpha \cdot MSE(f_{\theta}(X_r), Z_r)
\end{aligned}
\end{equation}

where task and replay cross-entropy are separately averaged over their respective unmasked next-response-token targets within each minibatch. The distillation term first averages mean squared error over response positions and vocabulary entries for each replay example, then averages over replay examples present. It compares the logits $f_{\theta}(X_r)$ of the model being trained (the student) against the pre-computed logits $Z_r$ of the frozen base model $M_0$ (the teacher). Canonical DER stores the model's logits when each sample enters the replay buffer \citep{buzzega_2020_dark}. Because our buffer is constructed before fine-tuning, these logits are generated by the base model, making $M_0$ the teacher.

We index the replay loss settings by these coefficients: \textbf{ER} ($\alpha{=}0, \beta{=}1$; cross-entropy replay only), \textbf{DER} ($\alpha{=}1, \beta{=}0$; logit distillation only) and \textbf{DER++} ($\alpha{=}1, \beta{=}1$; both). \textbf{Plain} uses task cross-entropy only with length-grouped task batches, giving 936 scheduled updates per stage. \textbf{Matched Plain} also has $\alpha{=}\beta{=}0$ but runs through the replay trainer: buffer rows are masked from task CE and both replay losses are zero, while those rows still occupy the mixed batches. Length grouping is therefore disabled and the 1032-update budget matches ER, DER and DER++. Matched Plain differs from the replay methods only in their loss terms and from original Plain only in batching and update budget. We evaluate it on both primary models, O1--O3 and seed 42.

To measure whether exposure and distillation have additive effects, we calculate their interaction separately on the ASR and ORR percentage-point scales. For each outcome, we calculate the interaction as $(\texttt{DER++}-\mathrm{DER})-(\mathrm{ER}-\mathrm{Plain})$ in percentage points. We also substitute Matched Plain for Plain to assess sensitivity to the task-only control.

\subsection{Evaluation Approach}
Each behavioral measure addresses a different failure mode. ASR measures harmful compliance, over-refusal rate (ORR) measures refusal of benign requests and task performance scores assess learning and retention of the downstream tasks.

HarmBench ASR is recorded across $M_0$--$M_3$ for the reported sequential safety trajectories; downstream task scores are also recorded across $M_0$--$M_3$, as forgetting calculations require complete chains for the relevant task. Held-out ASR and over-refusal are evaluated at $M_3$, with separate $M_0$ baselines, so the safety--over-refusal comparison is an endpoint analysis only. See Appendix~\ref{app:coverage} for a summary of the reported coverage.

\subsubsection{Task Performance}
Following TRACE \citep{wang_2023_trace}, we measure Py150 with edit similarity, ScienceQA with answer accuracy and 20Minuten with ROUGE-L.

\subsubsection{Attack Success Rate}
Attack Success Rate (ASR) is measured on the test split of the HarmBench DirectRequest dataset \citep{mazeika_2024_harmbench}. It contains 320 harmful prompts across categories such as copyright infringement, cybercrime, harassment/bullying and chemical/biological weaponry. Copyright infringement samples were discarded, mirroring \citet{arditi_2024_refusal}, so the reported metric covers the remaining categories ($n=240$).

The judge model, cais/HarmBench-Llama-2-13b-cls\footnote{https://huggingface.co/cais/HarmBench-Llama-2-13b-cls}, provides a binary harmful/safe label $\hat{y}_i$ for each generated output and is fine-tuned specifically to classify responses to the HarmBench dataset. ASR is the fraction of harmful prompts whose completion the judge labels harmful:
\begin{equation}
ASR = \dfrac{1}{N_{\text{harmful}}} \sum^{N_\text{harmful}}_{i=1} 1(\hat{y}_i = 1)
\end{equation}

To measure whether safety generalizes beyond HarmBench-style prompts, the final model $M_3$ is evaluated on three additional English-language jailbreak datasets (AdvBench \citep{zou_2023_universal}, JailbreakBench \citep{chao_2024_jailbreakbench} and the toxic split of OR-Bench \citep{cui_2025_orbench}), scored using the HarmBench judge model and ASR calculation above.

\subsubsection{Over-Refusal Rate}
Over-refusal is measured on benign prompts that resemble unsafe requests, commonly called hard negatives. WildGuard \citep{han_2024_wildguard} labels each prompt--completion pair as refusal or compliance. The over-refusal rate:
\begin{equation}
ORR = \dfrac{1}{N_{\text{safe}}} \sum^{N_\text{safe}}_{i=1} 1(\hat{y}_i = \text{refusal})
\end{equation}

is the fraction of hard-negative completions labeled as refusals; lower ORR means fewer benign requests are refused.

We evaluate on 500 randomly drawn samples from OR-Bench's \texttt{or-bench-hard-1k} subset of hard negative prompt completion pairs \citep{cui_2025_orbench}, labeled ORB-Hard in Section~\ref{sec:results} and the 250 hard negative samples in XSTest \citep{rttger_2024_xstest}.

\subsection{Analysis of Refusal Direction}
\label{sec:method-rq4-refusal}
We use difference-in-means (DIM) extraction and directional ablation \citep{arditi_2024_refusal} at each cumulative checkpoint to distinguish refusal geometry preservation from behavioral safety. DIM-vector magnitude measures the activation contrast between extraction classes, and cosine similarity measures angular drift from the base direction. Ablation then tests how much refusal depends on the checkpoint's own extracted direction. None of these measurements substitutes for the behavioral evaluations above.

The refusal analyses cover the $(\alpha, \beta)$ factorial from Section~\ref{sec:er_ft}: Plain, ER, DER and DER++ with the 70:30 harmful-to-benign replay buffer. For each method, we analyze geometry from the base model $M_0$ through the final model $M_3$, merging the LoRA adapters cumulatively.

\subsubsection{Extracting the refusal direction}
\label{sec:extract_refusal_direction}
Following \citet{arditi_2024_refusal}, we calculate the DIM vector from contrasting harmful and harmless prompts. $D_{\text{harmful}}$ contains OR-Bench-toxic samples and $D_{\text{harmless}}$ contains benign XSTest prompts; both sets are truncated to $n=112$.

For each checkpoint $M_k$ and layer $l$, we measure the residual-stream activation $\textbf{x}^{(l)}_{M_k}(t)$ at a fixed offset in the chat-formatted input: $-1$ for Mistral and $-5$ for Llama. The offsets are held fixed across checkpoints within each model. The refusal direction is the DIM vector between the two extraction classes (Equation~\ref{eq:diff_in_means}).

\begin{equation}
    \label{eq:diff_in_means}
    \textbf{r}^{(l)}_{M_k} = \frac{1}{|D_{\text{harmful}}|}\sum_{t \in D_{\text{harmful}}}\textbf{x}^{(l)}_{M_k}(t)
    - \frac{1}{|D_{\text{harmless}}|}\sum_{t \in D_{\text{harmless}}}\textbf{x}^{(l)}_{M_k}(t)
\end{equation}

The magnitude $\lVert\textbf{r}^{(l)}_{M_k}\rVert$ measures the extraction-set activation contrast at layer $l$. Its unit vector $\hat{\textbf{r}}^{(l)}_{M_k}=\textbf{r}^{(l)}_{M_k}/\lVert\textbf{r}^{(l)}_{M_k}\rVert$ is used for both cosine similarity to the base direction and the ablation in Section~\ref{sec:direction_ablation}. We use layer 12 for Llama, following \citet{arditi_2024_refusal}, and middle layer 16 for Mistral. The extraction layers remain fixed across checkpoints and have a measurable ablation effect at $M_0$. We assess class separation at $M_0$ and $M_3$ using in-sample AUROC: the same prompts are used to extract each direction and evaluate its projections, with AUROC averaged over O1--O3.

\subsubsection{Causal concentration via directional ablation}
\label{sec:direction_ablation}
Magnitude and cosine describe a single direction but do not establish its causal role in refusal. Thus, we remove the refusal unit vector $\hat{\textbf{r}}$ from every residual stream activation (Equation~\ref{eq:directional_ablation}) to prevent the model from representing this direction in the residual stream \citep{arditi_2024_refusal}.

\begin{equation}
    \label{eq:directional_ablation}
    \textbf{x}' \leftarrow \textbf{x} - \hat{\textbf{r}}\hat{\textbf{r}}^{\top}\textbf{x}
\end{equation}

For each checkpoint we generate unablated and ablated completions for 60 harmful OR-Bench-toxic prompts held out from the extraction set of Section~\ref{sec:extract_refusal_direction}; the ablated direction is never evaluated on prompts it was fit on. WildGuard \citep{han_2024_wildguard} judges each completion as refusal or compliance. Following \citet{arditi_2024_refusal}, $\text{RS}(M_k)$ denotes the refusal score of $M_k$ on harmful prompts and $\text{RS}(M_k \mid \hat{\mathbf{r}})$ the post-ablation score. Equation~\ref{eq:delta_refusal_ablation} gives the relative drop in refusal rate.

\begin{equation}
    \label{eq:delta_refusal_ablation}
    \Delta{\text{Refusal}} = 1 - \frac{\text{RS}(M_k\mid \hat{\textbf{r}})}{\text{RS}(M_k)}
\end{equation}

Magnitude $\lVert \textbf{r}^{(l)}_{M_k} \rVert$ and rotation $\cos(\hat{\textbf{r}}_{M_k}, \hat{\textbf{r}}_{M_0})$ \citep{du2025anchoring} describe how the direction changes during fine-tuning. A high $\Delta \text{Refusal}$ indicates that refusal is strongly mediated by the ablated direction.

\begin{figure*}[t]
    \centering
    \includegraphics[width=1\linewidth]{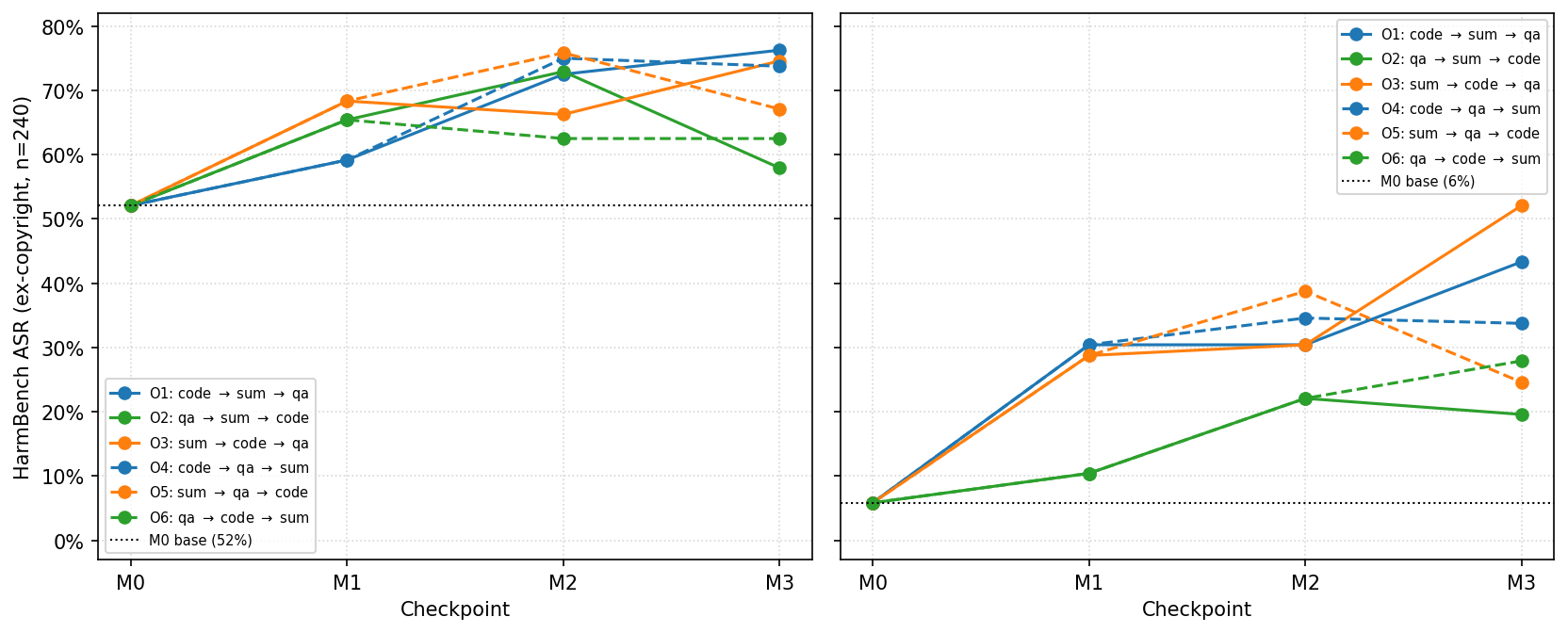}
    \caption{HarmBench DirectRequest ASR trajectories under Plain ($n{=}240$) for Mistral (left) and Llama (right), with orderings colored by first task and dashed $M_0$ baselines.}
    \label{fig:rq1_baseline_asr}
\end{figure*}

\subsection{SafeLoRA comparison}
\label{sec:projection_method}
SafeLoRA \citep{hsu_2024_safe} projects weight updates using aligned and unaligned model weights. We apply it post-hoc: the Plain training runs are completed first and projection is then applied after each cumulative merge of their saved adapters for O1--O3 at seed 42. No additional training is performed after projection. Consequently, each later adapter was learned from an unprojected predecessor, even though it is subsequently evaluated as part of a projected sequence. We use projection threshold $\tau=0.5$ on both primary models and examine threshold sensitivity on Mistral O1 by raising $\tau$ from $0.5$ to $1.0$; this paired rerun is reported separately from the O1--O3 averages and not pooled with them (see Table~\ref{tab:safelora_sensitivity}). Thresholds are not tuned per model and downstream task performance is not evaluated, so this is a sensitivity study rather than a tuned comparison with replay.

SafeLoRA \citep{hsu_2024_safe} pairs each instruction-tuned model with its \emph{unaligned} (base, pre-instruction-tuning) weights: \texttt{mistral-community/Mistral-7B-v0.2} for Mistral and \texttt{meta-llama/Meta-Llama-3-8B} for Llama. For each of 128 $q/k/v/o$ matrices, define $V_i=W_{\mathrm{aligned},i}-W_{\mathrm{unaligned},i}$ and $P_i=V_iV_i^\top/\lVert V_i\rVert_F$. These projection matrices remain fixed across tasks.

Saved Plain adapters are merged in task order. An update $\Delta W_{t,i}$ is replaced by $P_i\Delta W_{t,i}$ when their flattened cosine is at most $\tau$; otherwise it is retained. Adapters are not retrained after projection. The main comparison uses O1--O3, seed 42 and $\tau=0.5$.

\section{Results}
\label{sec:results}

Whether retaining a base model's behavior is desirable depends on a base model's safety operating point. Mistral-7B-Instruct-v0.2 has a HarmBench ASR of $52.1\%$, while Meta-Llama-3-8B-Instruct starts at $5.8\%$ ASR but refuses $72.4\%$ of benign OR-Bench-hard prompts. Staying close to base can therefore retain harmful compliance on Mistral or over-refusal on Llama. We assess both outcomes alongside task performance and refusal geometry.

Unless stated otherwise, factorial comparisons use final checkpoints averaged over the same three orderings, O1--O3, at seed 42. The wider Plain/DER++ comparison covers all six orderings, O1--O6.

\subsection{Task-only adaptation erodes safety in every ordering}
\label{sec:plain_ft_asr}

Every evaluated Plain sequence ends with higher HarmBench ASR than its base model, extending single-task evidence that benign fine-tuning can compromise safety \citep{qi_2024_finetuning, he_2024_whats, lermen_2024_lora}.  Averaged across all six orderings, ASR rises from $52.1\%$ to $68.7\%$ on Mistral and from $5.8\%$ to $33.5\%$ on Llama, as shown in Table~\ref{tab:rq1_asr}. Llama's mean increase is larger ($27.7$ versus $16.6$ percentage points). Safety erosion therefore affects both a model with substantial base harmful compliance and one with low initial ASR --- the catastrophic forgetting problem against which we evaluate replay.

\begin{table}[t]
\centering
\footnotesize
\setlength{\tabcolsep}{4pt}
\caption{HarmBench ASR (\%) under Plain across six task orderings and checkpoints $M_0$--$M_3$; $\Delta=M_3-M_0$.}
\label{tab:rq1_asr}
\begin{tabular}{llrrrrr}
\toprule
Model & Ordering & $M_0$ & $M_1$ & $M_2$ & $M_3$ & $\Delta$ \\
\midrule
\multirow{6}{*}{Mistral-7B} & O1 & 52.1 & 59.2 & 72.5 & 76.2 & $+24.2$ \\
 & O2 & 52.1 & 65.4 & 72.9 & 57.9 & $+5.8$ \\
 & O3 & 52.1 & 68.3 & 66.2 & 74.6 & $+22.5$ \\
 & O4 & 52.1 & 59.2 & 75.0 & 73.8 & $+21.7$ \\
 & O5 & 52.1 & 68.3 & 75.8 & 67.1 & $+15.0$ \\
 & O6 & 52.1 & 65.4 & 62.5 & 62.5 & $+10.4$ \\
\midrule
\multirow{6}{*}{Llama-3-8B} & O1 & 5.8 & 30.4 & 30.4 & 43.3 & $+37.5$ \\
 & O2 & 5.8 & 10.4 & 22.1 & 19.6 & $+13.8$ \\
 & O3 & 5.8 & 28.7 & 30.4 & 52.1 & $+46.2$ \\
 & O4 & 5.8 & 30.4 & 34.6 & 33.8 & $+27.9$ \\
 & O5 & 5.8 & 28.7 & 38.8 & 24.6 & $+18.8$ \\
 & O6 & 5.8 & 10.4 & 22.1 & 27.9 & $+22.1$ \\
\bottomrule
\end{tabular}
\end{table}

Task order determines the extent of that problem. Final ASR ranges from $57.9$--$76.2\%$ on Mistral and $19.6$--$52.1\%$ on Llama, with QA-last orderings ending highest. On Llama, QA raises ASR to only $10.4\%$ when trained first, compared with a mean of $47.7\%$ when trained last. Its effect therefore depends on the preceding adaptation. Erosion is not always monotonic: Mistral O2 ASR, for example, falls from $72.9\%$ at $M_2$ to $57.9\%$ at $M_3$; but no ordering restores the base level. Figure~\ref{fig:rq1_baseline_asr} shows the trajectories.

Seed replications repeat the erosion pattern (see Section~\ref{sec:seeds}), and homogeneous code-only chains erode safety as well, so task heterogeneity is not required for it, as shown in Table~\ref{tab:hom_control}. Here, $\Delta=M_3-M_0$ is in percentage points. These chains use 5000 examples versus 15{,}000 for heterogeneous chains, with fewer updates (see Appendix~\ref{app:training_budget}); final code edit similarity is approximately 53 points.

\begin{table}[ht]
\centering
\footnotesize
\setlength{\tabcolsep}{4pt}
\caption{HarmBench ASR (\%, $n{=}240$) over three Py150 shards at seed 42.}
\label{tab:hom_control}
\begin{tabular}{lrrrrr}
\toprule
Model & $M_0$ & $M_1$ & $M_2$ & $M_3$ & $\Delta$ \\
\midrule
Mistral-7B & 52.1 & 72.1 & 72.1 & 69.6 & $+17.5$ \\
Llama-3-8B & 5.8 & 34.2 & 37.9 & 28.7 & $+22.9$ \\
\bottomrule
\end{tabular}
\end{table}

\subsection{Exposure lowers ASR at an over-refusal cost; distillation stays near base}
\label{sec:method_decomp}

The factorial decomposition identifies two different responses to erosion. Supervised exposure (ER) drives harmful compliance well below base, while logit distillation (DER) largely retains base behavior. On Mistral, ER reduces mean HarmBench ASR to $1.0\%$, compared with $46.7\%$ for DER and $52.1\%$ at base, as shown in Table~\ref{tab:method_decomp}. DER therefore prevents much of the deterioration under Plain but leaves most of the base model's harmful compliance intact. Llama shows the same ordering, with less room to improve on its already low base ASR. The contrast also appears on the other harmful-request datasets (as shown later in Table~\ref{tab:method_decomp_full}).

\begin{figure}[ht]
    \centering
    \includegraphics[width=1\linewidth]{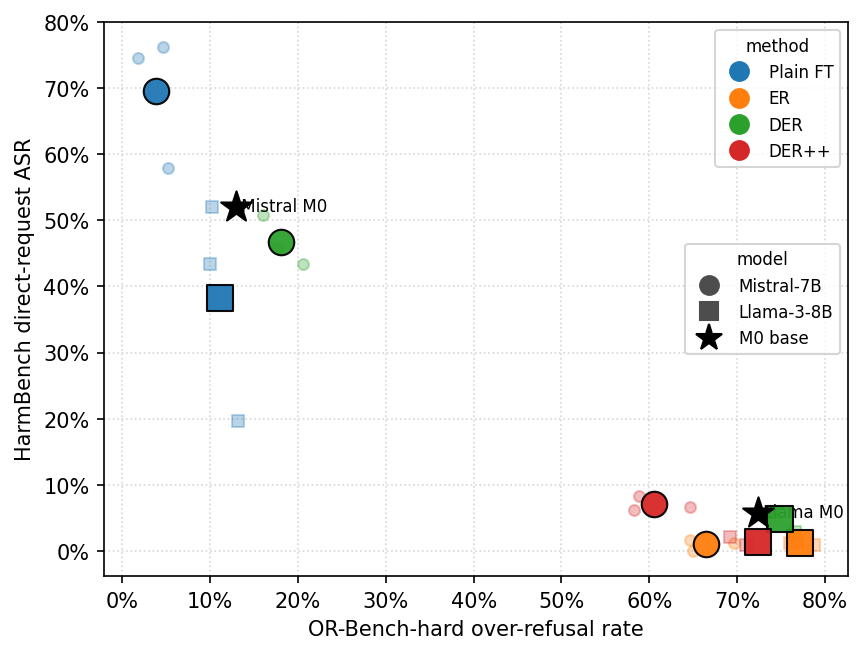}
    \caption{HarmBench ASR versus OR-Bench-hard ORR over O1--O3, seed 42 (lower is better on both axes). Faint points show orderings, bold points their means and stars the base models.}
    \label{fig:safety_overrefusal}
\end{figure}

Exposure's significant ASR reduction comes at the cost of a larger increase in benign refusal. On Mistral, OR-Bench-hard ORR rises from $13.0\%$ at base to $66.4\%$ under ER, compared with $18.1\%$ under DER. The same distinction appears on Llama, but its high base OR-Bench-hard ORR leaves less room for an increase. XSTest makes the cost clearer: ER raises Llama ORR from $4.8\%$ to $24.9\%$, while DER remains near base at $4.5\%$. Exposure's low ASR must therefore be interpreted alongside its greater tendency to refuse benign requests (Figure~\ref{fig:safety_overrefusal}).

Combining the replay components changes this trade-off. On Mistral, DER++ reduces ER's OR-Bench-hard ORR from $66.4\%$ to $60.5\%$, while ASR rises from $1.0\%$ to $7.1\%$. On Llama, DER++ also reduces ER's over-refusal on both benign benchmarks while retaining similarly low ASR. These effects are non-additive: the ASR reduction from adding exposure is smaller when distillation is already present, and adding distillation moderates exposure's over-refusal. Section~\ref{sec:interaction_checks} reports the interaction estimates and their sensitivity to the task-only control. No objective is uniformly preferable when considering both ASR and ORR.

Procedural and replication checks support this interpretation. Matching Plain to replay's batching and update budget shifts mean ASR by at most $3.3$ percentage points and ORR by at most $2.9$ points, well below the principal Plain--ER and Plain--DER++ gaps. Across three O1 seeds, the contrast between ER's low ASR/high over-refusal and DER's smaller deviation from base persists (as shown later in Table~\ref{tab:method_decomp_n3}). A single-ordering Gemma-2-9B-IT comparison \citep{google_2024_gemma} shows the same contrast, although its low base ASR limits the absolute difference between objectives (see Section~\ref{sec:gemma}).

\begin{table}[t]
\centering
\footnotesize
\setlength{\tabcolsep}{2.5pt}
\caption{Final-$M_3$ ASR and ORR (\%), mean $\pm$ sample s.d.\ over O1--O3 at seed 42.}
\label{tab:method_decomp}
\label{tab:balanced_factorial}
\begin{tabular}{llrrr}
\toprule
Model & Method & \shortstack{HarmBench\\ASR} & \shortstack{XSTest\\ORR} & \shortstack{ORB-hard\\ORR} \\
\midrule
\multirow{6}{*}{Mistral-7B} & $M_0$ & 52.1 & 4.8 & 13.0 \\
 & Plain & 69.6 $\pm$ 10.1 & 2.8 $\pm$ 0.7 & 3.9 $\pm$ 1.8 \\
 & Matched Plain & 71.2 $\pm$ 10.5 & 2.5 $\pm$ 1.2 & 4.9 $\pm$ 1.9 \\
 & ER & \textbf{1.0 $\pm$ 0.9} & 28.0 $\pm$ 2.8 & 66.4 $\pm$ 2.8 \\
 & DER & 46.7 $\pm$ 3.8 & 7.9 $\pm$ 0.8 & 18.1 $\pm$ 2.3 \\
 & DER++ & 7.1 $\pm$ 1.1 & 17.3 $\pm$ 0.5 & 60.5 $\pm$ 3.5 \\
\midrule
\multirow{6}{*}{Llama-3-8B} & $M_0$ & 5.8 & 4.8 & 72.4 \\
 & Plain & 38.3 $\pm$ 16.8 & 3.5 $\pm$ 2.0 & 11.1 $\pm$ 1.8 \\
 & Matched Plain & 41.7 $\pm$ 10.8 & 3.5 $\pm$ 2.7 & 8.3 $\pm$ 1.4 \\
 & ER & \textbf{1.2 $\pm$ 0.4} & 24.9 $\pm$ 2.7 & 77.2 $\pm$ 1.4 \\
 & DER & 4.9 $\pm$ 1.7 & 4.5 $\pm$ 0.6 & 74.9 $\pm$ 1.5 \\
 & DER++ & 1.4 $\pm$ 0.6 & 10.5 $\pm$ 1.6 & 72.3 $\pm$ 4.0 \\
\bottomrule
\end{tabular}
\end{table}

\begin{figure*}[t]
    \centering
    \includegraphics[width=0.95\textwidth]{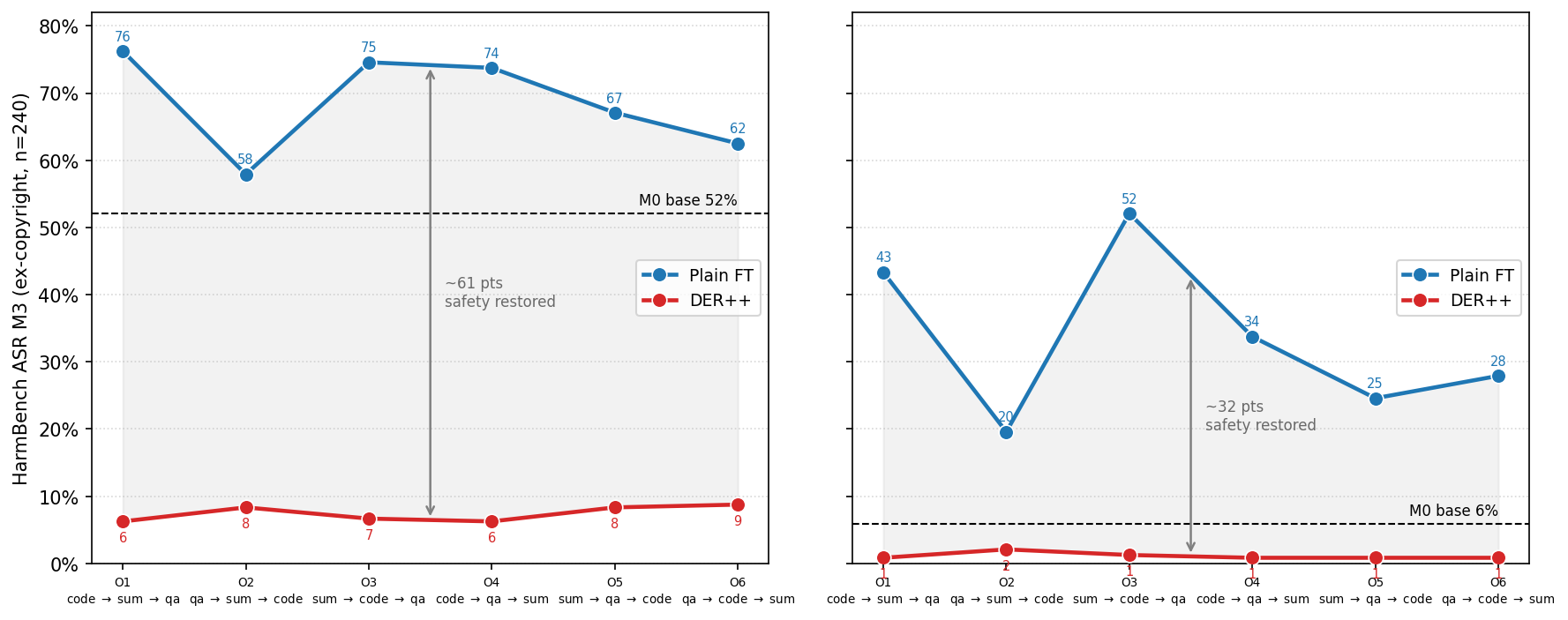}
    \caption{Final HarmBench ASR under Plain and DER++ across O1--O6 for Mistral (left) and Llama (right) ($n{=}240$, excluding copyright behaviors); dashed lines show $M_0$.}
    \label{fig:rq2_asr}
\end{figure*}

\subsubsection{Matched-control and interaction checks}
\label{sec:interaction_checks}

For the balanced O1--O3 factorial comparison, Matched Plain's mean ASR is $71.2\%$ versus $69.6\%$ for original Plain on Mistral, and $41.7\%$ versus $38.3\%$ on Llama. Calculated before rounding, the largest mean ASR and ORR shifts are $3.3$ and $2.9$ percentage points. These procedural effects are small relative to the Plain--ER and Plain--DER++ gaps, but do not establish equivalence between the controls.

On the percentage-point scales defined in Section~\ref{sec:er_ft}, the exposure--distillation interaction is $+29.0$ for Mistral HarmBench ASR and $+33.6$ for Llama; on XSTest ORR, it is $-15.7$ and $-15.5$. Using Matched Plain gives ASR interactions of $+30.6$ and $+37.0$, while XSTest ORR interactions change by at most $0.4$ points. These estimates describe non-additivity on the measured outcome scales, not an identified internal mechanism.

\begin{figure*}[ht]
    \centering
    \includegraphics[width=1\linewidth]{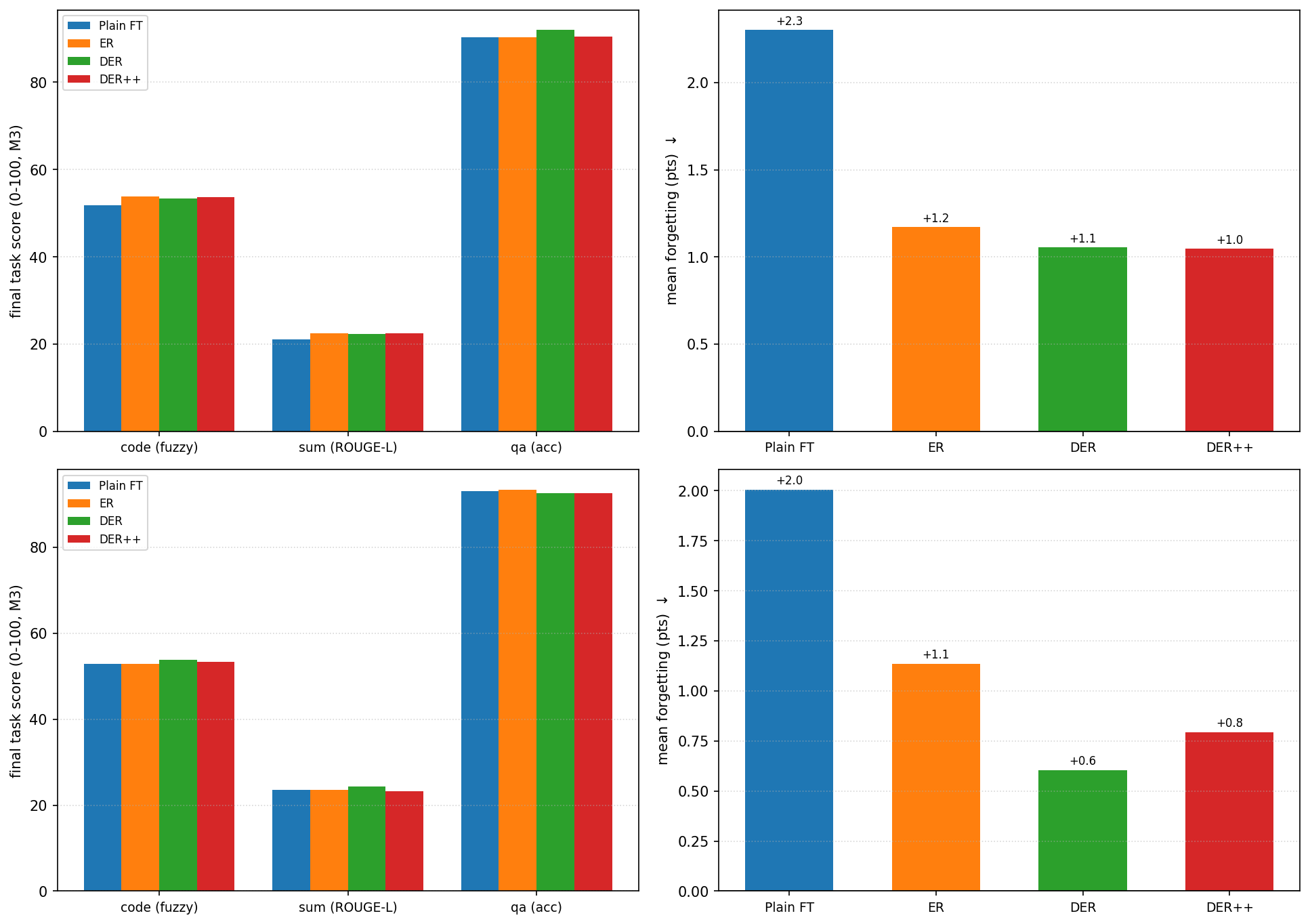}
    \caption{Downstream task performance at O3 $M_3$ (left) and mean forgetting over available orderings (right), with Mistral in the top row and Llama in the bottom row.}
    \label{fig:task_forgetting}
\end{figure*}

\subsection{DER++ retains task scores but still over-refuses}
\label{sec:der_pp_robust}
\label{sec:taskperf}

The combined objective's low ASR extends beyond the three-ordering factorial comparison. DER++ finishes below base in every evaluated ordering on both models (Figure~\ref{fig:rq2_asr}). Across all six orders, mean ASR is $7.4\%$ on Mistral and $1.1\%$ on Llama, compared with $68.7\%$ and $33.5\%$ under Plain. Task order changes the severity of unmitigated erosion, yet DER++ still reduces direct harmful compliance in every sequence.

\begin{table}[t]
\centering
\footnotesize
\setlength{\tabcolsep}{2pt}
\caption{Final-checkpoint $M_3$ task performance and forgetting, mean $\pm$ s.d.\ over available orderings with complete evaluation chains. Lowest mean forgetting per model in bold.}
\label{tab:taskperf}
\begin{tabular}{llrrrr}
\toprule
Model & Method & sum & code & qa & Forgetting \\
 & & (ROUGE-L) & (edit sim.) & (acc) & \\
\midrule
\multirow{5}{*}{Mistral-7B} & $M_0$ & 18.7 & 9.8 & 67.2 & --- \\
 & Plain & 21.4 $\pm$ 0.9 & 54.1 $\pm$ 1.2 & 87.0 $\pm$ 4.2 & 2.3 $\pm$ 1.9 \\
 & ER & 22.7 $\pm$ 0.5 & 53.2 $\pm$ 0.6 & 89.1 $\pm$ 1.8 & 1.2 $\pm$ 0.7 \\
 & DER & 22.1 $\pm$ 0.2 & 53.9 $\pm$ 0.6 & 90.8 $\pm$ 2.6 & 1.1 $\pm$ 1.3 \\
 & DER++ & 22.4 $\pm$ 0.6 & 53.6 $\pm$ 0.3 & 89.6 $\pm$ 1.3 & \textbf{1.0 $\pm$ 0.6} \\
\midrule
\multirow{5}{*}{Llama-3-8B} & $M_0$ & 18.6 & 7.3 & 72.5 & --- \\
 & Plain & 23.4 $\pm$ 0.9 & 52.8 $\pm$ 1.0 & 90.0 $\pm$ 2.9 & 2.0 $\pm$ 1.3 \\
 & ER & 23.4 $\pm$ 0.2 & 53.4 $\pm$ 0.9 & 92.0 $\pm$ 1.8 & 1.1 $\pm$ 0.6 \\
 & DER & 24.3 $\pm$ 0.1 & 53.9 $\pm$ 0.3 & 92.1 $\pm$ 1.0 & \textbf{0.6 $\pm$ 0.2} \\
 & DER++ & 23.7 $\pm$ 0.2 & 53.7 $\pm$ 0.3 & 92.2 $\pm$ 0.8 & 0.8 $\pm$ 0.1 \\
\bottomrule
\end{tabular}
\end{table}

\begin{table}[t]
\centering
\footnotesize
\setlength{\tabcolsep}{2pt}
\caption{DER++ final-$M_3$ HarmBench ASR and XSTest/ORB-hard ORR (\%) by ordering, with mean $\pm$ s.d.\ over O1--O6.}
\label{tab:rq2_asr}
\begin{tabular}{llrrr}
\toprule
Model & Ordering & HarmBench ASR & XSTest & ORB-hard \\
\midrule
\multirow{7}{*}{Mistral-7B} & O1 & 6.2 & 16.8 & 58.2 \\
 & O2 & 8.3 & 17.6 & 58.8 \\
 & O3 & 6.7 & 17.6 & 64.6 \\
 & O4 & 6.2 & 17.6 & 56.0 \\
 & O5 & 8.3 & 17.6 & 63.8 \\
 & O6 & 8.8 & 22.4 & 67.8 \\
 & \textbf{Mean $\pm$ s.d.} & 7.4 $\pm$ 1.2 & 18.3 $\pm$ 2.1 & 61.5 $\pm$ 4.5 \\
\midrule
\multirow{7}{*}{Llama-3-8B} & O1 & 0.8 & 10.8 & 71.0 \\
 & O2 & 2.1 & 8.8 & 69.2 \\
 & O3 & 1.2 & 12.0 & 76.8 \\
 & O4 & 0.8 & 10.0 & 66.6 \\
 & O5 & 0.8 & 11.2 & 74.8 \\
 & O6 & 0.8 & 9.6 & 75.2 \\
 & \textbf{Mean $\pm$ s.d.} & 1.1 $\pm$ 0.5 & 10.4 $\pm$ 1.2 & 72.3 $\pm$ 4.0 \\
\bottomrule
\end{tabular}
\end{table}

The task evaluations show that continued learning of the downstream tasks survives this reduction. As shown in Table~\ref{tab:taskperf}, scores are similar across Plain and the replay objectives; for example, Mistral's mean final code edit similarity spans only $53.2$--$54.1$, and summarization and QA scores remain above base. Task performance and forgetting are averaged over available orderings. Forgetting averages the peak-to-final task score drop on the two tasks not trained last. Replay also shows lower observed mean forgetting on earlier tasks than Plain (Figure~\ref{fig:task_forgetting}), at most $1.2$ points compared with $2.3$ points for Plain.

Those task scores leave the benign-refusal cost largely unmeasured. Across all six orderings, Mistral DER++ refuses $61.5\%$ of OR-Bench-hard prompts, compared with $13.0\%$ at base. Llama stays near its already high base OR-Bench-hard ORR, but its XSTest ORR rises from $4.8\%$ to $10.4\%$, as shown in Table~\ref{tab:rq2_asr}. Thus, similar downstream scores cannot establish that ordinary benign use is unaffected; task performance and over-refusal must be measured separately.

\begin{table*}[t]
\centering
\footnotesize
\setlength{\tabcolsep}{6pt}
\caption{Final-$M_3$ ASR and ORR (\%) at seed 42, mean $\pm$ s.d.\ over O1--O6 for Plain/DER++ and O1--O3 for ER/DER.}
\label{tab:method_decomp_full}
\begin{tabular}{llrrrrr}
\toprule
Model & Metric & $M_0$ & Plain & DER & ER & DER++ \\
\midrule
\multirow{6}{*}{Mistral-7B} & HarmBench ASR & 52.1 & 68.7 $\pm$ 7.4 & 46.7 $\pm$ 3.8 & 1.0 $\pm$ 0.9 & 7.4 $\pm$ 1.2 \\
 & AdvBench ASR & 35.8 & 66.3 $\pm$ 5.6 & 30.3 $\pm$ 1.8 & 0.0 $\pm$ 0.0 & 0.7 $\pm$ 0.4 \\
 & JBB ASR & 41.0 & 73.5 $\pm$ 8.4 & 39.0 $\pm$ 4.4 & 0.3 $\pm$ 0.6 & 1.0 $\pm$ 0.6 \\
 & ORB-toxic ASR & 43.1 & 75.8 $\pm$ 2.8 & 37.4 $\pm$ 2.4 & 2.0 $\pm$ 0.5 & 8.3 $\pm$ 2.5 \\
 & XSTest ORR & 4.8 & 2.7 $\pm$ 0.6 & 7.9 $\pm$ 0.8 & 28.0 $\pm$ 2.8 & 18.3 $\pm$ 2.1 \\
 & ORB-hard ORR & 13.0 & 4.0 $\pm$ 1.6 & 18.1 $\pm$ 2.3 & 66.4 $\pm$ 2.8 & 61.5 $\pm$ 4.5 \\
\midrule
\multirow{6}{*}{Llama-3-8B} & HarmBench ASR & 5.8 & 33.5 $\pm$ 12.2 & 4.9 $\pm$ 1.7 & 1.2 $\pm$ 0.4 & 1.1 $\pm$ 0.5 \\
 & AdvBench ASR & 1.2 & 12.8 $\pm$ 4.6 & 1.1 $\pm$ 0.3 & 0.0 $\pm$ 0.0 & 0.0 $\pm$ 0.0 \\
 & JBB ASR & 3.0 & 25.3 $\pm$ 10.5 & 2.3 $\pm$ 1.5 & 0.3 $\pm$ 0.6 & 0.2 $\pm$ 0.4 \\
 & ORB-toxic ASR & 4.1 & 28.6 $\pm$ 5.1 & 2.9 $\pm$ 0.7 & 2.5 $\pm$ 0.5 & 4.7 $\pm$ 0.8 \\
 & XSTest ORR & 4.8 & 4.1 $\pm$ 2.0 & 4.5 $\pm$ 0.6 & 24.9 $\pm$ 2.7 & 10.4 $\pm$ 1.2 \\
 & ORB-hard ORR & 72.4 & 12.3 $\pm$ 3.4 & 74.9 $\pm$ 1.5 & 77.2 $\pm$ 1.4 & 72.3 $\pm$ 4.0 \\
\bottomrule
\end{tabular}
\end{table*}

\subsection{Additional safety results and buffer composition}
\label{sec:extended_safety} 

Additional harmful-request and over-refusal benchmarks are also analyzed (Table~\ref{tab:method_decomp_full}). Unlike the balanced comparison in Table~\ref{tab:method_decomp}, its means use different ordering sets: O1--O6 for Plain and DER++, and O1--O3 for ER and DER.

On Llama, however, DER++ has mean OR-Bench-toxic ASR of $4.7\%$, compared with $4.1\%$ at base, so its below-base HarmBench results do not extend to every benchmark.

The available buffer comparison also changes the trade-off. On Mistral O2 and O3, the 50:50 harmful:hard-negative buffer has 11--15 points lower OR-Bench-hard ORR and 1--6 points higher ASR than the main 70:30 buffer (Table~\ref{tab:buffer_comp}). This comparison keeps the buffer at 500 examples while changing its harmful-benign ratio. The two evaluated orderings suggest that adding benign replay examples can moderate over-refusal, with some loss of the harmful-compliance reduction. However, coverage is limited to Mistral O2/O3 so it does not identify a generally optimal buffer ratio.

\begin{table}[ht]
\centering
\footnotesize
\setlength{\tabcolsep}{4pt}
\caption{Final-checkpoint $M_3$ ASR and ORR (\%) for the 70:30 and 50:50 buffers on Mistral O2/O3.}
\label{tab:buffer_comp}
\begin{tabular}{lrrrr}
\toprule
Metric (Mistral-7B) & O2 70:30 & O2 50:50 & O3 70:30 & O3 50:50 \\
\midrule
HarmBench ASR & 8.3 & 13.8 & 6.7 & 7.9 \\
XSTest ORR & 17.6 & 16.0 & 17.6 & 16.8 \\
ORB-hard ORR & 58.8 & 48.0 & 64.6 & 50.0 \\
\bottomrule
\end{tabular}
\end{table}

\subsection{Lower ASR does not require preservation of the base refusal direction}
\label{sec:refusal_tracking}

\begin{figure}[ht]
    \centering
    \includegraphics[width=1\linewidth]{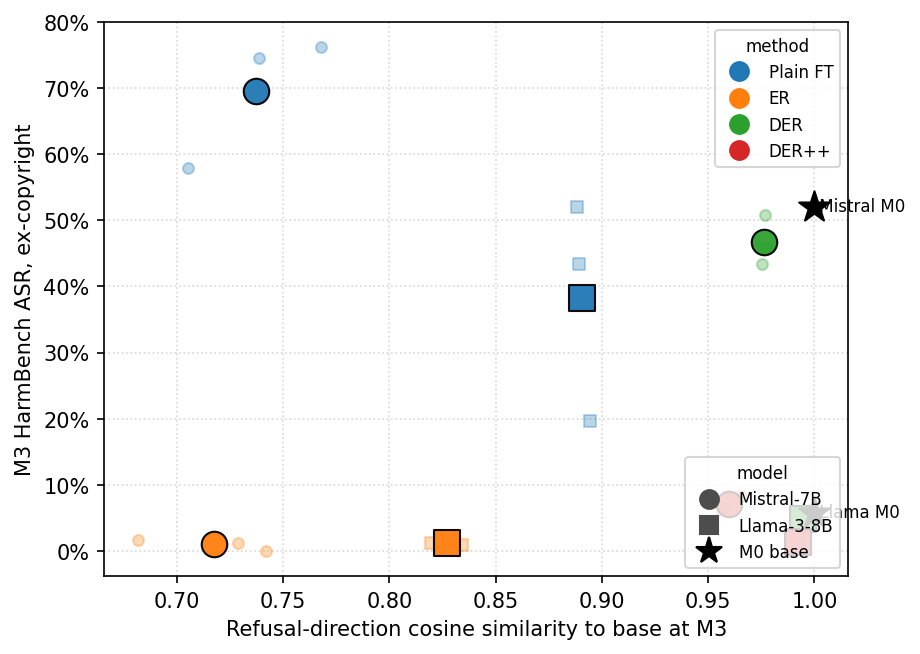}
    \caption{Final-checkpoint cosine similarity to the base refusal direction versus HarmBench ASR (faint: orderings, bold: means, stars: $M_0$).}
    \label{fig:decoupling}
\end{figure}

Refusal direction tracking demonstrates that retaining the base representation does not identify the training objective with the lowest harmful compliance. On both models, DER preserves the direction more closely than ER, despite ER achieving lower ASR (Figure~\ref{fig:decoupling}). The mismatch is particularly clear for Mistral: DER's cosine similarity to base is $0.976$, versus $0.718$ for ER, while their ASRs are $46.7\%$ and $1.0\%$.

\begin{figure*}[t]
    \centering
    \includegraphics[width=1\linewidth]{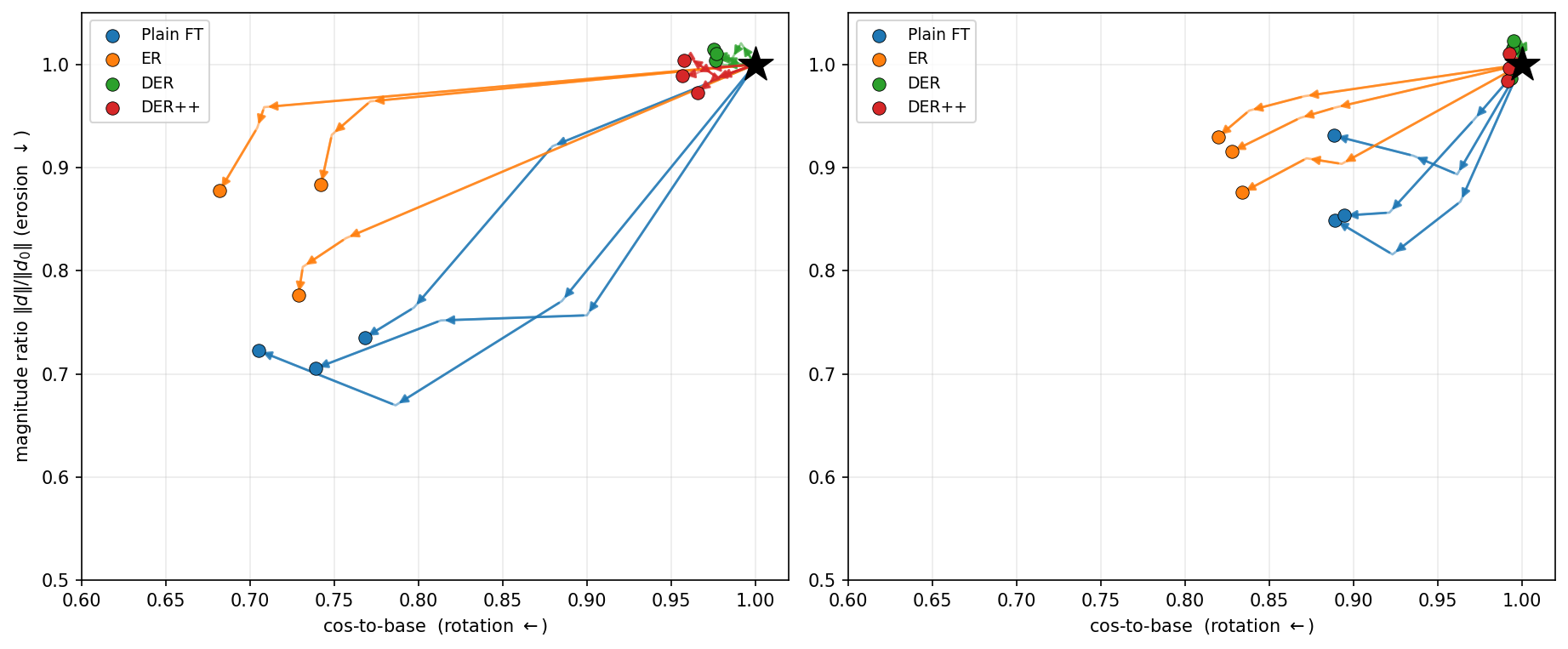}
    \caption{Refusal geometry across O1--O3 for Mistral (left) and Llama (right), with magnitude normalized to each trajectory's base and each arrow representing one task step.}
    \label{fig:rq4_mag_cos}
\end{figure*}

\begin{figure}[ht]
    \centering
    \includegraphics[width=1\linewidth]{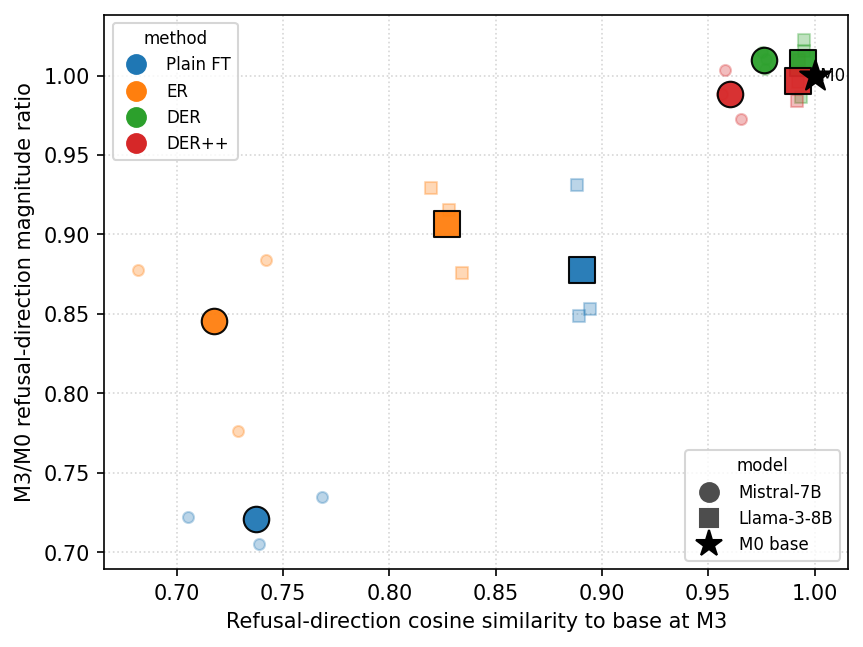}
    \caption{$M_3$ cosine similarity to the base refusal direction versus relative magnitude over O1--O3 (faint: orderings; bold: model--method means).}

    \label{fig:erosion_rotation}
\end{figure}

The wider geometric pattern follows the training signal. DER and DER++ retain high cosine similarity and nearly unchanged magnitude on both models, while Plain and ER rotate farther from base (Table~\ref{tab:refusal_geom}; Figure~\ref{fig:rq4_mag_cos}). Since Plain increases ASR and ER reduces it, rotation alone cannot distinguish safety erosion from the behavioral change induced by supervised replay exposure. See Figure~\ref{fig:erosion_rotation} for a summary of the magnitude and rotation of the refusal direction across the objectives.

Direction similarity alone cannot tell us whether the extracted direction still mediates refusal. In Table~\ref{tab:refusal_geom}, Refusal is the unablated refusal rate and $\Delta$Refusal is the fraction of WildGuard-judged refusal removed by own-direction ablation on held-out prompts (Eq.~\ref{eq:delta_refusal_ablation}); a low unablated refusal rate magnifies this ratio. On Llama, ablating ER's own direction at $M_3$ removes only $6\%$ of refusals, compared with $71\%$ for DER; this gap persists across O1 seeds (as described later in Table~\ref{tab:refusal_geom_n3}). On Mistral, ER again has a smaller effect than DER but the gap is narrower ($0.19$ versus $0.32$ in $\Delta$Refusal).

Mistral Plain illustrates why $\Delta$Refusal must be read as a relative change. Its unablated refusal rate at $M_3$ is only $0.15$, and its $\Delta$Refusal of $-0.29$ means refusal increases after ablation. Because the unablated rate is the denominator in Eq.~\ref{eq:delta_refusal_ablation}, $-0.29$ does not mean a 29-percentage-point increase. A supplementary Gemma evaluation likewise finds a smaller effect for ER than DER, though the DER--DER++ ordering differs from Llama (see Section~\ref{sec:gemma}). How much refusal survives directional ablation is therefore model-dependent, yet across all three models geometric preservation still fails to rank the observed ASR outcomes.

\begin{table}[ht]
\centering
\footnotesize
\setlength{\tabcolsep}{3pt}
\caption{Refusal geometry and ablation effects, means over O1--O3 at $M_3$, except the $\Delta$Refusal chain ($M_1{\rightarrow}M_2{\rightarrow}M_3$). Negative $\Delta$Refusal indicates increased refusal after ablation; no cells are bolded as these metrics have no preferred direction.}
\label{tab:refusal_geom}
\begin{tabular}{llrrrl}
\toprule
Model & Method & $\cos$ & $\lVert \textbf{r} \rVert$ & Refusal & $\Delta$Refusal \\
\midrule
\multirow{5}{*}{Mistral-7B} & $M_0$ & 1.000 & 2.89 & 0.72 & 0.42 \\
 & Plain & 0.737 & 2.08 & 0.15 & $0.15\rightarrow-0.16\rightarrow-0.29$ \\
 & ER & 0.718 & 2.44 & 0.99 & $0.16\rightarrow0.24\rightarrow0.19$ \\
 & DER & 0.976 & 2.92 & 0.70 & $0.33\rightarrow0.28\rightarrow0.32$ \\
 & DER++ & 0.960 & 2.85 & 0.94 & $0.28\rightarrow0.31\rightarrow0.30$ \\
\midrule
\multirow{5}{*}{Llama-3-8B} & $M_0$ & 1.000 & 2.77 & 0.97 & 0.69 \\
 & Plain & 0.891 & 2.43 & 0.64 & $0.69\rightarrow0.65\rightarrow0.65$ \\
 & ER & 0.827 & 2.51 & 1.00 & $0.03\rightarrow0.05\rightarrow0.06$ \\
 & DER & 0.995 & 2.79 & 0.98 & $0.71\rightarrow0.72\rightarrow0.71$ \\
 & DER++ & 0.992 & 2.76 & 0.95 & $0.61\rightarrow0.57\rightarrow0.54$ \\
\bottomrule
\end{tabular}
\end{table}

Figure~\ref{fig:erosion_rotation} compares final refusal-direction similarity and magnitude across O1--O3. Magnitudes are normalized to the base values of 2.89 for Mistral and 2.77 for Llama.

\subsection{Seed robustness}
\label{sec:seeds}

Seed replications cover O1, seeds 42--44, excluding Matched Plain. Table~\ref{tab:method_decomp_n3} reports ASR and ORR. Table~\ref{tab:refusal_geom_n3} reports $\Delta$Refusal, the fraction of refusals removed by own-direction ablation (Eq.~\ref{eq:delta_refusal_ablation}). O1 Plain endpoints are $75.3\pm1.3\%$ on Mistral and $41.0\pm2.5\%$ on Llama. Across the four objectives, ASR standard deviations range from $0.5$ to $2.5$ percentage points, smaller than the main gaps between Plain and exposure-based replay. At seed 43 on Mistral, final-task mean ASR retains the ordering QA ($74.8\%$), summarization ($70.6\%$), then code ($66.9\%$).

\begin{table}[ht]
\centering
\footnotesize
\setlength{\tabcolsep}{4pt}
\caption{O1 final-$M_3$ ASR and ORR (\%), mean $\pm$ s.d.\ over seeds 42--44. Lowest mean per measure and model in bold.}
\label{tab:method_decomp_n3}
\begin{tabular}{llrrr}
\toprule
Model & Method & \shortstack{HarmBench\\ASR} & \shortstack{XSTest\\ORR} & \shortstack{ORB-hard\\ORR} \\
\midrule
\multirow{5}{*}{Mistral-7B} & $M_0$ & 52.1 & 4.8 & 13.0 \\
 & Plain & 75.3 $\pm$ 1.3 & \textbf{2.9 $\pm$ 0.5} & \textbf{5.9 $\pm$ 1.3} \\
 & ER & \textbf{1.1 $\pm$ 1.0} & 27.5 $\pm$ 2.0 & 67.7 $\pm$ 3.8 \\
 & DER & 45.4 $\pm$ 1.8 & 7.1 $\pm$ 0.6 & 18.5 $\pm$ 2.0 \\
 & DER++ & 8.3 $\pm$ 2.1 & 17.9 $\pm$ 1.2 & 57.7 $\pm$ 3.8 \\
\midrule
\multirow{5}{*}{Llama-3-8B} & $M_0$ & 5.8 & 4.8 & 72.4 \\
 & Plain & 41.0 $\pm$ 2.5 & \textbf{2.9 $\pm$ 0.5} & \textbf{10.9 $\pm$ 1.9} \\
 & ER & 1.5 $\pm$ 0.6 & 26.5 $\pm$ 1.6 & 83.8 $\pm$ 4.7 \\
 & DER & 4.2 $\pm$ 2.2 & 5.3 $\pm$ 0.6 & 72.5 $\pm$ 4.6 \\
 & DER++ & \textbf{0.6 $\pm$ 0.5} & 11.2 $\pm$ 0.4 & 72.9 $\pm$ 2.1 \\
\bottomrule
\end{tabular}
\end{table}

\begin{table}[ht]
\centering
\footnotesize
\setlength{\tabcolsep}{4pt}
\caption{O1 final-$M_3$ $\Delta$Refusal, mean $\pm$ s.d.\ over seeds 42--44.}
\label{tab:refusal_geom_n3}
\begin{tabular}{llr}
\toprule
Model & Method & $\Delta$Refusal \\
\midrule
\multirow{5}{*}{Mistral-7B} & $M_0$ & 0.42 \\
 & Plain & $-$0.26 $\pm$ 0.36 \\
 & ER & 0.20 $\pm$ 0.03 \\
 & DER & 0.28 $\pm$ 0.03 \\
 & DER++ & 0.27 $\pm$ 0.04 \\
\midrule
\multirow{5}{*}{Llama-3-8B} & $M_0$ & 0.69 \\
 & Plain & 0.65 $\pm$ 0.08 \\
 & ER & 0.06 $\pm$ 0.01 \\
 & DER & 0.69 $\pm$ 0.02 \\
 & DER++ & 0.58 $\pm$ 0.08 \\
\bottomrule
\end{tabular}
\end{table}

\subsection{Gemma generalization check}
\label{sec:gemma}

Gemma-2-9B-IT uses seed 42, token offset $-5$, extraction layer 21, micro-batch 4 and gradient accumulation 4 (effective batch size 16), with the same 60 ablation prompts as the primary models. Across O1--O6, Plain raises HarmBench ASR from $2.1\%$ to $5.8$--$15.4\%$ (mean $8.5\%$); DER++ ends at $1.7$--$2.5\%$ (mean $2.1\%$). Table~\ref{tab:gemma_factorial} gives the O1 behavioral results. ER reduces ASR below base ($1.2\%$ versus $2.1\%$), while DER retains the base value; their OR-Bench-hard ORRs are $91.6\%$ and $82.4\%$, compared with $79.0\%$ at base. This single-ordering comparison reproduces the distinction between exposure's larger ASR reduction and distillation's retention of base behavior.

ER has a smaller own-direction ablation effect than DER ($0.15$ versus $0.59$). DER++ however has a larger effect than DER ($0.64$ versus $0.59$), reversing their ordering on Llama. The ER--DER contrast recurs with Gemma but the full ordering of geometric effects does not generalize across models.

\begin{table}[ht]
\centering
\footnotesize
\setlength{\tabcolsep}{5pt}
\caption{Gemma O1 factorial ASR and ORR at $M_3$ (\%).}
\label{tab:gemma_factorial}
\begin{tabular}{lrrr}
\toprule
Method & HarmBench ASR & XSTest ORR & ORB-hard ORR \\
\midrule
$M_0$ & 2.1 & \textbf{16.0} & 79.0 \\
Plain & 15.4 & 18.0 & \textbf{60.6} \\
ER & \textbf{1.2} & 26.8 & 91.6 \\
DER & 2.1 & 17.2 & 82.4 \\
DER++ & 2.5 & 16.8 & 78.6 \\
\bottomrule
\end{tabular}
\end{table}

\subsection{SafeLoRA also exhibits a safety--over-refusal trade-off}
\label{sec:projection_defense}

Post-hoc SafeLoRA reduces harmful compliance on Llama while increasing refusal of benign requests. Thus the safety--over-refusal trade-off observed under replay also occurs under weight-space projection in this setting. At the shared threshold $\tau=0.5$, Llama's mean HarmBench ASR over O1--O3 falls from $38.3\%$ under Plain to $20.3\%$, while OR-Bench-hard ORR rises from $11.1\%$ to $43.7\%$. ASR remains above the base model's $5.8\%$. Lower harmful compliance thus comes with a substantial benign-refusal cost under the post-hoc protocol in Section~\ref{sec:projection_method}.

The ASR effect varies with the model and projection threshold. At $\tau=0.5$, Mistral's mean ASR is $72.4\%$, above both Plain ($69.6\%$) and base ($52.1\%$). In a separate Mistral O1 rerun, however, raising $\tau$ from $0.5$ to $1.0$ lowers ASR from $77.5\%$ to $50.4\%$, near base (see Table~\ref{tab:safelora_sensitivity}). This threshold sensitivity limits conclusions from the shared setting.

Table~\ref{tab:safelora_sensitivity} gives the Mistral O1 threshold sweep. ASR is the percentage of harmful answers on 240 matched non-copyright HarmBench behaviors; the base model is shown for context, not as a threshold setting. Projecting the ten lowest-cosine matrices instead gives mean O1--O3 ASR of $74.7\%$ (Mistral) and $36.5\%$ (Llama). Sensitivity runs round selection cosines to five decimal places; their $\tau=0.5$ means are $72.1\%$/$20.8\%$, versus $72.4\%$/$20.3\%$ in the original evaluations above. These runs are not pooled.

\begin{table}[ht]
\centering
\footnotesize
\setlength{\tabcolsep}{8pt}
\caption{SafeLoRA threshold sensitivity on Mistral O1 at seed 42.}
\label{tab:safelora_sensitivity}
\begin{tabular}{lrr}
\toprule
Setting & Harmful / total & ASR (\%) \\
\midrule
$M_0$ & 125/240 & 52.1 \\
$\tau=0.1$ & 186/240 & 77.5 \\
$\tau=0.3$ & 175/240 & 72.9 \\
$\tau=0.5$ & 186/240 & 77.5 \\
$\tau=1.0$ & 121/240 & 50.4 \\
\bottomrule
\end{tabular}
\end{table}

\section{Discussion}
\label{sec:discussion}

\subsection{Retaining base behavior and reducing harmful compliance are different objectives}
The factorial decomposition shows why the choice of replay signal matters. Distillation constrains the model toward the base logits on the buffer, so its behavior changes little. Supervised exposure instead trains on the buffer's refusal and compliance responses, producing larger reductions in ASR along with more benign refusal. These effects follow directly from what each objective optimizes and, depending on the base model's safety profile, can be desirable or detrimental.

For Mistral, this proximity reproduces substantial harmful compliance; for Llama, it retains substantial over-refusal of difficult benign requests. Adding distillation to exposure moderates over-refusal but sacrifices some of exposure's ASR reduction on Mistral. These non-additive effects make the relative cost of harmful compliance and benign refusal, rather than any single optimal setup, the important choice when selecting a training objective.

\subsection{Refusal geometry and behavioral safety decouple}
The geometric results reinforce the distinction between direction preservation and ASR reduction. Distillation preserves the base direction closely on both primary models but exposure reaches lower ASR after greater rotation. Plain also rotates the direction while increasing ASR. Table~\ref{tab:factorial_summary} summarizes this mismatch: similar degrees of geometric preservation can result in very different behavioral outcomes. Direction tracking is therefore informative about representational changes but is insufficient on its own to select a safety-preserving objective.

Llama's ablation result adds a distinct limitation. After supervised exposure, the extracted direction separates the extraction prompts yet accounts for little of the observed refusal under intervention. This does not establish that refusal has become multidimensional: a different direction or extraction layer could account for the retained behavior. Distinguishing those possibilities requires layer sweeps and multi-directional interventions \citep{wollschlger_2025_the, faaizjoad_2026_there}.

\begin{table}[ht]
\centering
\footnotesize
\renewcommand{\arraystretch}{1.0}
\setlength{\tabcolsep}{6pt}
\caption{Descriptive $(\alpha,\beta)$ factorial summary; $\alpha$: distillation, $\beta$: replay exposure, cosine: similarity to the base refusal direction, ASR: HarmBench.}
\label{tab:factorial_summary}
\begin{tabular}{lcc}
\toprule
 & $\beta=0$ & $\beta=1$ \\
\midrule
\multirow{3}{*}{$\alpha=0$}
 & \textbf{Plain} & \textbf{ER} \\
 & Lower base cosine & Lower base cosine \\
 & ASR above base & ASR below base \\
\midrule
\multirow{3}{*}{$\alpha=1$}
 & \textbf{DER} & \textbf{DER++} \\
 & Higher base cosine & Higher base cosine \\
 & ASR near base & ASR below base \\
\bottomrule
\end{tabular}
\end{table}

\subsection{Safety evaluation requires complementary behavioral tests}
\label{sec:projection_discussion}
Task performance and benign refusal expose different aspects of the safety-overrefusal trade-off. Similar task scores show that the replay objectives retain learning on the evaluated tasks, but do not reveal how often the model refuses difficult benign requests. A mitigation can therefore preserve task learning and reduce harmful compliance while remaining costly for benign use. Evaluating all three outcomes is necessary to characterize that cost.

The Llama SafeLoRA result extends the need for joint behavioral evaluation beyond replay: weight-space projection also reduces harmful compliance while increasing benign refusal. Its Mistral threshold sensitivity and unmeasured task utility prevent a general comparison with replay. What generalizes across these evaluations is the need to assess each behavioral outcome directly.

\section{Limitations and Future Work}
\label{sec:limitations}
The main study covers two 7--8B instruction-tuned models and three-task sequences, with a single-ordering 9B Gemma check. Whether these trade-offs extend to longer chains, larger models and reasoning models remains untested.

The reported safety evaluations use fixed harmful requests and do not establish robustness to adaptive attacks. Future work should evaluate optimized suffix attacks such as GCG \citep{zou_2023_universal}, as well as multi-turn, transfer and prompt-injection attacks, to test whether the exposure--distillation trade-off persists under stronger adversarial pressure.

All ASR and ORR values come from automated judges, without a human audit of errors. Directional ablation uses one extraction layer per model and 60 WildGuard-scored prompts from a single dataset; its effects may depend on that layer, dataset and judge. Our geometric analysis does not fully characterize how refusal is represented. Larger ablation datasets, layer sweeps and multi-directional refusal analyses would enable a more complete understanding.

The factorial comparison holds several design choices fixed. ER, DER and DER++ share the same procedure and budget, while Matched Plain checks the difference from original Plain only on O1--O3 at seed 42. Its small observed shifts do not establish equivalence across other orders or seeds. Adapter rank and replay-buffer size are fixed at 16 and 500 examples, with no sensitivity ablations, and the buffer comes from one source, WildJailbreak.

Controlled sweeps of adapter rank, replay volume, buffer construction and loss coefficients would establish whether the same distinction between exposure and distillation persists beyond these settings.

\section{Conclusion}
\label{sec:conclusion}

Sequential fine-tuning erodes safety across every evaluated task ordering, but the two replay signals address that erosion differently. Supervised exposure reduces harmful compliance well below base at a substantial benign-refusal cost; logit distillation largely retains base behavior. Combining them moderates exposure's over-refusal while retaining task learning, with some loss of its ASR reduction on Mistral. The geometric analysis supports the same distinction between retention and behavioral improvement: preserving the base refusal direction does not identify the lowest-ASR objective, and Llama's refusal after exposure largely survives ablation of its extracted direction. These findings make the replay objective consequential for both harmful and benign use. Choosing it requires direct evaluation of harmful compliance and benign refusal alongside task performance.

\section*{Open Science}
The repository includes the results supporting the reported evaluations, training and evaluation code, offline analysis scripts and figure assets. The results folders contain retained aggregate metrics and per-item records, including prompts, generated responses and judge outputs. Model weights and raw training/replay inputs are omitted because of their size. The release includes the dataset and replay-buffer builders and the replay seed to support inspection of the construction procedure. Base model identifiers are listed in Appendix~\ref{app:coverage} and training datasets in Appendix~\ref{app:impl}. The repository is available at \url{https://github.com/gruellan/preserving_safety_sequential_finetuning}.

\section*{LLM usage considerations}
LLM assistants supported coding, statistical analysis and paper revisions. LLMs were used for editorial purposes in this manuscript, and all outputs were inspected by the authors to ensure accuracy and originality. The substantial majority of the code and writing was performed by the authors. The authors verified all methods, calculations and commentary outputted by the LLMs and take full responsibility for the final work.

LLMs are also part of the experimental method: $M_0$ provides replay-filtering responses and teacher logits used in fine-tuning experiments, HarmBench supplies the harmfulness classifier model and WildGuard supplies refusal labels. These labels are not human annotations and their error rates on the evaluated conditions are unknown. The study uses 7--9B models, LoRA, cached teacher logits and reused saved outputs to make comparisons computationally feasible on a single A100 and minimize redundant computation. Restricting seed replication reduces compute while also limiting generalization.

\section*{Ethical Considerations}
This study evaluates safety degradation and replay mitigations using public models and established benchmarks. The paper presents aggregate metrics, without reproducing harmful prompts or completions. The release includes the results supporting the reported evaluations, including retained prompts and model responses. These records support scrutiny of the evaluations but may contain harmful or disturbing material, and releasing them carries a minor risk of misuse.

\clearpage

\printbibliography

\clearpage

\section{Appendix}

\subsection{Implementation details}
\label{app:impl}

LoRA targets the $q/k/v/o$ projections with rank 16, scaling 32 and dropout $0.05$. Training uses 3 epochs per task, effective batch size 16, peak learning rate $2\times10^{-4}$, AdamW (weight decay $0.01$) and cosine decay with 3\% warmup. Python, NumPy and PyTorch are seeded (42 unless stated); deterministic kernels are used. Primary models use SDPA, Gemma uses eager attention, on a single NVIDIA A100 SXM. Each TRACE \texttt{LLM-CL-Benchmark\_5000} split supplies 5000 examples (Py150, 20Minuten, ScienceQA), totaling 15{,}000 per heterogeneous ordering. The homogeneous control uses three disjoint Py150 shards (1667/1667/1666 examples).

\subsection{Batching and training budgets}
\label{app:training_budget}

Section~\ref{sec:er_ft} defines the mixed batches, loss normalization and matched control. Both collators pad dynamically and mask prompt and padding labels with $-100$. Input limits, including responses, are 1024 tokens for code, QA and replay, and 2048 for summarization. Empty task or replay groups contribute zero loss; gradient accumulation combines minibatch losses rather than a global token-weighted mean. The trainer skips logit comparison when $\alpha=0$.

Original Plain has 936 scheduled updates per task; Matched Plain and replay have 1032. The homogeneous control has 312 per stage (936 total), versus 2808 for heterogeneous Plain. 

\subsection{Evaluation coverage and retained records}
\label{app:coverage}

Table~\ref{tab:coverage} lists the reported run sets. ASR uses 240 non-copyright HarmBench, 520 AdvBench, 100 JailbreakBench and 320 OR-Bench-toxic requests. ORR uses 250 benign XSTest and 500 OR-Bench-hard prompts; unsafe XSTest prompts are excluded. Geometry uses 112 prompts per extraction class and 60 held-out ablation prompts.

\begin{table}[ht]
\centering
\footnotesize
\setlength{\tabcolsep}{3pt}
\caption{Reported evaluation coverage by measurement, training condition and model checkpoint.}
\label{tab:coverage}
\begin{tabular}{p{0.27\columnwidth}p{0.34\columnwidth}p{0.29\columnwidth}}
\toprule
Measurement & Conditions & Checkpoints \\
\midrule
Primary HarmBench & Plain/DER++ O1--O6; ER/DER O1--O3; seed 42 & $M_0$--$M_3$ \\
Primary held-out ASR and ORR & Same method/ordering sets & $M_0$, $M_3$ \\
Matched Plain control & Primary models, O1--O3, seed 42 & $M_0$, $M_3$ ASR/ORR \\
Primary geometry/ablation & Four methods, O1--O3, seed 42 & $M_0$--$M_3$ \\
Primary seed replication & Four methods, O1, seeds 42--44 & ASR/geometry chains; held-out ASR/ORR endpoints \\
Gemma check & Plain/DER++ O1--O6; four methods at O1; seed 42 & ASR chains and ORR endpoints; geometry O1--O3 for Plain/DER++ and O1 for ER/DER \\
\bottomrule
\end{tabular}
\end{table}

Model identifiers are \path{mistralai/Mistral-7B-Instruct-v0.2}, \path{meta-llama/Meta-Llama-3-8B-Instruct} and \path{google/gemma-2-9b-it}.

\end{document}